\documentclass{article}
\usepackage{ijcai22}

\usepackage{times}
\usepackage{soul}
\usepackage{url}
\usepackage[hidelinks]{hyperref}
\usepackage[utf8]{inputenc}
\usepackage[small]{caption}
\usepackage{graphicx}
\usepackage{amsmath}
\usepackage{amsthm}
\usepackage{booktabs}
\usepackage{algorithm}
\usepackage{algorithm,algorithmicx,algpseudocode}
\usepackage{booktabs,subcaption,amsfonts,dcolumn}
\usepackage {tikz}
\usetikzlibrary {positioning}
\usepackage[subnum]{cases}

\usepackage{balance}

\usepackage{wrapfig}

\usepackage{caption}
\usepackage[separate-uncertainty=true]{siunitx}
\usepackage{mathtools}
\usepackage{graphicx}
\usepackage{courier}
\usepackage{epstopdf}
\usepackage{color}
\usepackage{csquotes}

\usepackage{amsmath}
\usepackage{amsfonts}
\usepackage{amssymb}
\usepackage[subnum]{cases}
\usepackage{booktabs} %For formal tables
\usepackage{pifont}

\usepackage{algorithm,algorithmicx,algpseudocode}

\usepackage{amssymb}% http://ctan.org/pkg/amssymb
\usepackage{pifont}% http://ctan.org/pkg/pifont

\usepackage {tikz}
\usetikzlibrary {positioning}
\definecolor {processblue}{cmyk}{0.96,0,0,0}
\usepackage{lipsum,adjustbox}

\usepackage[subnum]{cases}
\usepackage{color,soul}

\usepackage{subcaption}

\usepackage{footnote}
\makesavenoteenv{tabular} %ICDE version does not support

\usepackage{multirow}

\usepackage{booktabs,subcaption,amsfonts,dcolumn}
\newcolumntype{d}[1]{D..{#1}}
\usepackage{xcolor}

\usepackage{textcomp}

\title{Dynamic Structure Learning through Graph Neural Network for\\ Forecasting Soil Moisture in Precision Agriculture}

\author{
Anoushka Vyas$^1$
\and
Sambaran Bandyopadhyay$^2$\footnote{The work was done when Anoushka was an intern at IBM Research and Sambaran was affiliated to IBM Research prior to joining Amazon}\footnote{Both Anoushka and Sambaran contributed equally}
\affiliations
$^1$Department of Computer Science, Virginia Tech, Arlington, VA, USA\\
$^2$Amazon
\emails
anoushkav@vt.edu,
samb.bandyo@gmail.com
}

\begin{document}

\maketitle

\begin{abstract}
    Soil moisture is an important component of precision agriculture as it directly impacts the growth and quality of vegetation. Forecasting soil moisture is essential to schedule the irrigation and optimize the use of water. Physics based soil moisture models need rich features and heavy computation which is not scalable. In recent literature, conventional machine learning models have been applied for this problem. These models are fast and simple, but they often fail to capture the spatio-temporal correlation that soil moisture exhibits over a region. In this work, we propose a novel graph neural network based solution that learns temporal graph structures and forecast soil moisture in an end-to-end framework. Our solution is able to handle the problem of missing ground truth soil moisture which is common in practice. We show the merit of our algorithm on real-world soil moisture data.
\end{abstract}

\section{Introduction}\label{sec:intro}
Precision agriculture \cite{zhang2002precision} is the science of observing, assessing and controlling agricultural practices such as monitoring soil, crop and climate in a field; detection and prevention of pest and disease attacks; providing a decision support system. It can help optimizing the natural resources needed for agricultural activities and thus ensures an efficient, sustainable and environment friendly development of agriculture. Remote sensing and IoT technology \cite{liaghat2010review} can be used as effective tools in precision agriculture by providing high resolution satellite imagery data with rich information about crop status, crop and water stresses, and ground truth agricultural information from local sensors and agricultural drones. Thus, availability of historical and real-time data has significantly improved the scope of applying artificial intelligence for precision agricultural practices \cite{jha2019comprehensive}. %,smith2020getting}.

Soil moisture is an important component of precision agriculture \cite{zhang2002precision} for crop health and stress management, irrigation scheduling, food quality and supply chain. Soil moisture measures the amount of water stored in various depths of soil. Forecasting high resolution and accurate soil moisture well ahead of time helps to save natural resources such as water. Soil moisture can be measured by soil sensors in the field \cite{hummel2001soil} or by physics based land surface models \cite{rui2011readme}. However, deploying soil moisture sensors on a vast region is expensive. They also cannot provide forecasts. Physics based soil moisture estimation can be quite accurate, but they need extremely rich set of input features such as different soil properties, landscape information, crop information which are difficult to obtain. These physical models are also computationally heavy, making them infeasible to scale over a larger region.

Soil moisture is also directly related to weather parameters such as temperature, precipitation, and vegetation condition of the field such as Normalized Difference Vegetation Index (NDVI) \cite{pettorelli2013normalized}. There are data driven and machine learning approaches present for soil moisture estimation. Conventional techniques such as Bayesian estimation, random forest and support vector regression are commonly used in industry for their simplicity to model soil moisture with respect to these flexible set of input features \cite{dasgupta2019soil,ahmad2010estimating}. However, these approaches essentially treat different locations independently while modeling soil moisture and fail to exploit the rich spatial dependency that soil moisture values over a region exhibit. For example, if there is a rainfall, soil moisture values over the whole region increases. In an agricultural region, soil moisture values exhibits similar patterns with varying crop cycles.
There exist methods in machine learning to capture such spatial dependencies through latent variable models \cite{castro2012latent} and supervised deep learning architectures \cite{zhang2017deep}. But they are complex in nature (with large number of parameters) and difficult to apply for soil moisture estimation as getting ground truth soil moisture data is expensive because of the cost of deploying physical sensors.

In this paper, we propose soil moisture forecasting as a semi-supervised learning problem on dynamic graphs (also known as temporal graphs or sequence of graphs) where the structure of the graph and node attributes change over time to capture the spatio-temporal variation of soil moisture.
In contrast to existing supervised techniques, please note that semi-supervised nature of our solution enables us to train the network on both labeled (data points with ground truth soil moisture) and unlabeled (data points with missing soil moisture) data. This is important because (i) obtaining ground truth soil moisture is expensive and (ii) there can be large amount of missing ground truth soil moisture values because of device and communication failures.
A graph can be visualized as a connected (by edges) set of entities (nodes). Intuitively, more the correlation between the soil moisture values in two locations more should be the chance of them to be connected in the graph. Recently, graph representation learning, or more specifically graph neural networks (GNNs) \cite{wu2019comprehensive,kipf2017semi,bandyopadhyay2021unsupervised} are able to achieve state-of-the-art performance on several learning tasks on graph. GNNs are also proposed to operate on dynamic graphs for traffic forecasting \cite{li2018diffusion,DBLP:conf/aaai/ChenCXCGF20} and social networks \cite{pareja2020evolvegcn}.
However, these GNN based approaches cannot be used directly for modeling soil moisture. Unlike social networks where the graph structure is explicitly given \cite{karrer2011stochastic}, there is no ground truth graph structure given as an input for the problem of spatio-temporal data modeling. As discussed before, it is possible that the correlation between soil moisture values in two nearby locations is negligible due to multiple hidden factors. Thus, we also need to learn the graph structure of the problem along with the prediction of soil moisture. Recently, research is conducted on graph structure learning using GNNs \cite{zhu2021deep}. However, those approaches are often limited to graphs without any temporal aspect and parameterized over the entries of the discrete adjacency matrix which make the optimization computationally expensive. As discussed in Section \ref{sec:graphLearning}, we tackle these problems in a simple and effective way by introducing multiple regularizers. Following are the novel contributions we make in this paper:
\begin{itemize}
    \item We pose the problem of soil moisture forecasting as a semi-supervised learning problem on dynamic graphs to jointly capture the varying degrees of spatial and temporal dependencies. As no ground truth graph structure is given as an input, we learn and update the sequence of graph structures in an end-to-end fashion.
    \item We propose a novel dynamic graph neural network, which is referred as DGLR (\underline{D}ynamic \underline{G}raph \underline{L}earning through \underline{R}ecurrent graph neural network).
    \item We curate two real world soil moisture datasets. The source code of DGLR and the datasets are available at \url{https://github.com/AnoushkaVyas/DGLR}.
\end{itemize}

\section{Problem Formulation}\label{sec:prob}
We are given a set of $N$ locations indexed by the set $[N] = \{1,2,\cdots,N\}$ from a geographic region. There are $[T] = \{1,2,\cdots,T\}$ time steps. For each time step $t$, ground truth soil moisture (SM) values are given for a subset of locations $N_{tr}^t \subseteq [N]$. Please note that $N_{tr}^{t1}$ may be different from $N_{tr}^{t2}$ for $t1 \neq t2$ because of the presence of missing SM values.

We use $s_i^t \in \mathbb{R}$ to denote the soil moisture at location $i$ and time step $t$. For each location $i$ and time step $t$, there are some $D$ input features (such as temperature, relative humidity, precipitation, NDVI, etc.) available which might be useful to predict soil moisture. Let us denote those feature by an attribute vector $x_i^t \in \mathbb{R}^D$, $\forall i \in [N], t \in [T]$. We also assume that the geographic distance between any two locations are given. Let us use $d_{ij}$ to denote the distance between two locations $i$ and $j$. Given all these information, our goal is to predict (forecast) the soil moisture at time step $T+1$ for each location $i \in [N]$, such that the forecasting model considers both past soil moisture and input feature values, and also able to learn and exploit the relation between different locations in their soil moisture values.

In practice, retraining the model for forecasting soil moisture for every time step in future is expensive. So for our experiments in Section \ref{sec:exp}, we train (including validation) the models roughly on the first 70\% to 80\% of the time interval and predict soil moisture on the rest.

\section{Solution Approach}\label{sec:soln}
There are multiple components of our integrated solution DGLR. Please note that all the parameters are trained in an end-to-end fashion. We explain each component below.

\subsection{Initial Dynamic Graph Formulation}\label{sec:initial}
As explained in Section \ref{sec:intro}, there is no explicit graph structure given for the problem of soil moisture forecasting. To apply graph neural network in the first iteration, we use the following heuristic to form an initial graph structure. First, for each location $i \in [N]$, we form a node (indexed by $i$) in the graph. Intuitively, if two locations are very close by, there soil moisture values can be more correlated compared to the points which are far away (there are exceptions to this and our learning algorithm is going to handle those). So, we connect any two nodes by an undirected and unweighted edge in the graph if the distance between the two corresponding locations $d_{ij}$ is less than some pre-defined threshold $\theta$. We also create a self-loop for every node in the graph. We assign attribute vector $x_i^t \in \mathbb{R}^F$ (as discussed in Section \ref{sec:prob}) to node $i$ at time $t$. According to this construction, the link structure of the graph is same across different time steps, but node features are changing. Thus, the constructed graph is dynamic in nature\footnote{A dynamic or temporal graph is a sequence of (correlated) graphs over time.}. Let us denote the set of graphs as $\{G^1,\cdots,G^T\}$, where $G^t = (V, E^t, X^t)$ is the graph\footnote{Initially, the edge set is same for all the graphs, but they will change during the course of learning.} at time step $t$ and $V=[N]$. We denote the adjacency matrix of $G^t$ as $A^t$. We also row-normalize the adjacency matrix by dividing each row with the degree of the corresponding node.

\subsection{Temporal Graph Neural Network}\label{sec:GNN}
Given a dynamic graph, we develop a temporal graph neural network which can generate embedding of a node in each time step and also use that to forecast soil moisture. Each layer of the proposed temporal graph neural network has two major components. There is a self-attention based GNN (with shared parameters) similar to \cite{velickovic2018graph} which works on each graph, thus capturing the spatial dependency between the nodes. The updated node embeddings from the GNN is fed to a RNN which connects graphs over different time steps to capture the temporal dependency of soil moisture. The layer is formally discussed below.

For a graph $G^t$, we use a trainable parameter matrix $W$ (shared over graphs $\forall t \in [T]$) to transform the initial feature vector $x_i^t$ as $W x_i^t$. As different neighboring locations may have significantly different impact on the soil moisture of a location, we use a trainable attention vector $a \in \mathbb{R}^{2K}$ to learn the importance $\alpha_{ij}$ for any two neighboring nodes in a graph as follows:
\begin{flalign}
    \alpha_{ij}^t = \frac{exp\Big( \text{LeakyReLU} \Big( a \cdot [W x_i^t || W x_j^t] \Big) \Big)}{ \sum\limits_{j' \in \mathcal{N}_{G^t}(i)} exp\Big( \text{LeakyReLU} \Big( a \cdot [W x_i^t || W x_{j'}^t] \Big) \Big) }
\end{flalign}
where $\mathcal{N}_{G^t}(i)$ is the neighboring nodes of $i$ (including $i$ itself) in the graph $G^t$, $\cdot$ represents dot product between the two vectors and $||$ is the vector concatenation. These normalized importance parameters are used to update the node features as:
\begin{flalign}\label{eq:GNNemb}
    h_i^t = \sigma \Big( \sum\limits_{j \in \mathcal{N}_{G^t}(i)}\alpha_{ij}^t A_{ij}^t W x_j^t\Big)
\end{flalign}
where $A_{ij}^t$ is the $(i,j)$the element of $A^t$, i.e., weight of the edge $(i,j)$ in $G^t$. 

The sequence of updated node embeddings $h_i^t$ of a node $i$ over different time steps are fed to a Gated Recurrent Unit (GRU) \cite{chung2014empirical}, which is a popularly used recurrent neural network. The GRU unit for $i$th node at time step $t$ takes the GNN output $h_i^t$ (from Equation \ref{eq:GNNemb}) and the GRU output $h_i^{t-1}$ to update $h_i^t$, as shown below\footnote{We use the same notation $h_i^t$ as the output of both GNN and GRU. But they update it sequentially.}.
\begin{flalign}
    U_i^t=\sigma(W_U^i h_i^{t-1}+P_U^i h_i^{t-1}+B_U^i); \\
    R_i^t= \sigma(W_R^i h_i^t+P_R^i h_i^{t-1}+B_R^i); \\
    \Tilde{h}_i^t = \text{tanh}(W_H^i h_i^t+P_H^i( R_i^t \circ h_i^{t-1} )+B_H^i); \\
    h_i^t = (1- U_i^t) \circ h_i^{t-1} + U_i^t \circ  \Tilde{h}_i^t
\end{flalign}
where are update and reset gates of GRU at time $t$, respectively. $W_U^i,W_R^i,W_H^i,P_U^i,P_R^i,P_H^i,B_U^i,B_R^i,B_H^i$ are trainable parameters of GRU for the node $i$. Please note that we have not shared the parameters of GRUs for different nodes. This helps to boost the performance as the temporal trend of soil moisture values are quite different in different locations.
The above completes the description of a GNN+GRU layer of our temporal graph neural network. The output of the GRU of a layer is fed as input to the next layer of GNN. Two such layers of GNN-GRU pair are used for all our experiments. We denote $H^t = [h_1^t \cdots h_N^t]^{Trans}$, $\forall t \in [T]$.

For soil moisture prediction at time $t$, node embeddings from the final layer GRU are used. The node embeddings from $t-w$ to $t-1$ are used where $w$ is a window of time steps. We concatenate $h_i^{t-w},\cdots,h_i^{t-1}$ and pass the vector to a fully connected layer with ReLU activation to predict the soil moisture of $i^{th}$ node at time $t$, which is denoted as $\hat{s}_i^t$. The window $w$ is a sliding window with stride 1. The loss of soil moisture prediction is calculated as:
\begin{flalign}\label{eq:STSM}
    \mathcal{L}_{STSM} = \sum\limits_{t=w+1}^T \sum\limits_{i \in N_{tr}^t} (s_i^t - \hat{s}_i^t)^2 
\end{flalign}

\subsection{Updating the Dynamic Graph Structures}\label{sec:graphLearning}
In Section \ref{sec:initial}, we use a simple heuristic to form the initial graph which may not always reflect the actual spatial dependency of soil moisture values. In this section, we try to learn the link structure between different nodes to improve the quality of the graph such that it facilitates soil moisture prediction.
Graph structure learning has received attention in recent literature. But a wide varieties of approaches are only limited to static graphs without temporal aspects and parameterize all the entries of the adjacency matrix (i.e., learning edges between the nodes) which makes the optimization combinatorial in nature \cite{fatemi2021slaps,selvan2020graph,shang2020discrete}. Our approach is quite feasible as we learn the temporal graph structures via node embeddings by introducing simple regularizers as discussed below.

First, we reconstruct the adjacency matrix $\hat{A}^t$ of the graph $G^t$ by the similarity of the node embeddings. Formally, we obtain $\hat{A}^t$ from the node embedding matrix $H^t$ as:
\begin{flalign}\label{eq:recon}
    \hat{A}^t = \text{ReLU} \Big( H^t {H^t}^{Trans} \Big) \in \mathbb{R}^{N \times N}
\end{flalign}
Thus, $A_{ij}^t$ (weight of an edge) is the dot product of the two corresponding node embeddings $h_i^t = H_{i:}^t$ and $h_j^t = H_{j:}^t$. The element-wise use of activation function ReLU$(\cdot)$ ensures that reconstructed edge weights are non-negative and thus can be used in the message passing framework of GNN. Similar to $A^t$, we also row-normalize $\hat{A}^t$ by degrees of the nodes.
However, the reconstructed graph can be noisy if the initially obtained node embeddings are erroneous. To avoid that, we use following regularization terms.

\textbf{Graph Closeness}:
Initially constructed graph with adjacency structure $A^t$, $\forall t \in [T]$, though not perfect, is easy to explain and intuitive as soil moisture values in close by regions tend to be correlated. So we want the reconstructed graph $\hat{A}^t$ not to move very far away from $A^t$. To ensure that, we use the following graph closeness regularization:
\begin{flalign}\label{eq:GC}
    \mathcal{L}_{GC} = \sum\limits_{t=1}^T \text{Distance}(A^t, \hat{A^t})
\end{flalign}
One can use any distance function between the matrices such as KL divergence or any matrix norms. We use binary cross entropy between $A^t$ and $\hat{A^t}$ as that produces better results.

\textbf{Feature Smoothness}: As explained in Section \ref{sec:intro}, features like weather parameters, vegetation type of the field have significant impact on soil moisture. So, locations having similar features often exhibit similar soil moisture values. It is not necessary that these locations have to be close to each other through geographic distance. Thus, we aim to connect such locations in the link structure of the graphs by using the feature smoothness regularizer as shown below:
\begin{flalign}\label{eq:FS}
    \mathcal{L}_{FS} = \sum\limits_{t=1}^T \sum\limits_{\substack{i,j \in [N] \\ i \neq j}} \hat{A}_{ij}^t||x_i^t - x_j^t||_2^2
\end{flalign}
Minimizing above ensures that for any time step $t \in [T]$, if feature vectors for two nodes $i$ and $j$ are similar, i.e., $||x_i^t - x_j^t||_2^2$ is less, the optimization would try to assign higher values to $\hat{A}_{ij}^t$. Similarly, if $||x_i^t - x_j^t||_2^2$ is high for two locations $i$ and $j$ at time step $t$, feature smoothness would try to lower the weight of the edge $(i,j)$ even if their geographic distance $d_{ij}$ is low. 

\textbf{Target Smoothness}: Finally, if two nodes have similar target variable (which is soil moisture for this work), message passing between them improves the performance of graph neural network, as observed in \cite{Hou2020Measuring}. For example in a graph dataset, if nodes which are directly connected by an edge also share same node labels, it helps the performance of node classification via message passing GNNs. As soil moisture is continuous in nature, we use the square of the difference of two soil moisture values to find their similarity.
We calculate this regularization term only on the training set $N_{tr}^t$ for each time step $t$, where ground truth soil moisture is available. 
\begin{flalign}\label{eq:TS}
    \mathcal{L}_{TS} = \sum\limits_{t=1}^T \sum\limits_{\substack{i,j \in N_{tr}^t \\ i \neq j}} \hat{A}_{ij}^t(s_i^t - s_j^t)^2
\end{flalign}
Again, minimizing target smoothness ensures that less edge weights are assigned to a node pair where nodes have very different soil moisture values. Thus, during the application of GNN, it avoids mixing features which lead to different types of soil moisture values.

\subsection{Joint Optimization and Training}
There are multiple loss components in the overall solution of DGLR. So, we form the final loss function of DGLR by taking a linear combination of all as shown below.
\begin{flalign}\label{eq:totalLoss}
    \underset{W_{G}, \Theta_{R}}{\text{min}} \mathcal{L}_{total} = \alpha_1\mathcal{L}_{STSM} + \alpha_2 \mathcal{L}_{GC} + \alpha_3 \mathcal{L}_{FS} + \alpha_4 \mathcal{L}_{TS}
\end{flalign}
where $\alpha_1, \alpha_2, \alpha_3, \alpha_4 \in \mathbb{R}_+$ are hyperparameters. $W_{G}$ and $\Theta_{R}$ contain the trainable parameters of GNN and RNN layers respectively. In contrast to existing works which solve graph structure learning as an expensive approximation of a combinatorial optimization problem \cite{jin2020graph,selvan2020graph}, the variables in the optimization problem in Equation \ref{eq:totalLoss} are only the parameters of GNN and RNNs. Please note that we do not minimize Equations \ref{eq:GC}-\ref{eq:TS} directly with respect to the reconstructed adjacency matrix $\hat{A}$, rather it is always calculated as $\hat{A}^t = \text{ReLU} \Big( H^t {H^t}^{Trans}\Big)$ (Equation \ref{eq:recon}) (followed by row normalisation). Thus, instead of solving task learning (which in this case is predicting soil moisture) and graph structure learning as alternating minimization, we jointly solve both in the framework of neural networks using ADAM \cite{kingma2014adam}. 

Once we obtain the reconstructed adjacency matrix $\hat{A}^t$ after optimizing Equation \ref{eq:totalLoss}, we use $\hat{A}^t$, along with node feature matrices $X^t$, $\forall t \in [T]$ for the next iteration of DGLR. The pseudo code of DGLR is presented in Algorithm \ref{alg:DGLR}. 

Please note that DGLR is semi-supervised in nature. For example, there can be examples in the training where soil moisture value is not present due to some loss of information or faulty sensors, but other feature information is present. In a conventional supervised setup, such unlabeled examples (without ground truth) are ignored in the training. But in DGLR, both labeled and unlabeled examples are used to propagate information through the graph neural networks (Equation \ref{eq:GNNemb}). Unlabeled examples can also contribute in the training loss function components such as Equations \ref{eq:GC} and \ref{eq:FS}. Because of the semi-supervised nature, DGLR can work well even with lesser amount of labeled data or missing data, as shown in Section \ref{sec:sensiSM}. 

Computation of the GNNs takes $O(MTDK)$ time, where $M$ is the average number of edges in the initial and reconstructed graphs over time. GRU takes another $O(NTK)$ time to generate embeddings. Finally, the loss components collectively takes $O(N^2TK)$ time. Hence, each iteration of DGLR takes $O((N^2+MD)TK)$ time. One can reduce the runtime of graph update regularizers by adding restriction on two nodes to get connected by an edge if their physical distance is more than some threshold, particularly for larger datasets.
Besides, forming a graph across locations which are very far from each other does not make sense because of the significant change of input features, landscape, soil types, etc. Hence, DGLR can be used for real-world soil moisture prediction.

\begin{algorithm} %[H]%[tb]
  \small
  \caption{\textbf{DGLR}}
  %\resizebox{0.47\textwidth}{!} {
  \label{alg:DGLR}
%\resizebox{0.5\textwidth}{!}{\begin{minipage}{\textwidth}   
\begin{algorithmic}[1]
      
	\Statex \textbf{Input}: Soil moisture locations $[N]$ and distances $d_{ij}$, $\forall i,j \in [N]$ between them. Temporal attribute matrix $X^t = \{ x_i^t \; | \; \forall i \in [N] \}$ for each time step $t \in [T+L]$. Ground truth soil moisture values for (a subset of) locations from time $1,\cdots,T$. Window length $w$.
    \Statex \textbf{Output}: Predict soil moisture values of all the locations from $T+1, \cdots, T+L$.
	\Statex \textbf{Training}
	\State Construct an initial attributed graph $G^t$ with normalized adjacency matrix $A^t$ and node attribute matrix $X^t$ for each time $t \in [T]$ where each node represents one location (Section \ref{sec:initial}).
	\State Set $\Tilde{A}^t = A^t$, $\forall t \in [T]$
	\For{pre-defined number of iterations}
    	\State Pass $\Tilde{A}^t$ and $X^t$ to two layers of GNN+RNN pairs (Section \ref{sec:GNN}) to generate all the node embeddings, $\forall t \in [T]$.
    	\State Concatenate $h_i^{t-w},\cdots,h_i^{t-1}$ and pass it to a fully connected layer to predict soil moisture $\hat{s}_i^t$, $\forall i$ and $\forall t \in \{w+1,\cdots,T\}$
		\State Update the parameters of GNN and RNNs by minimizing the losses at Equation \ref{eq:totalLoss}.
		\State Obtain $\hat{A^t}$ by Equation \ref{eq:recon}, followed by row normalization and set $\Tilde{A}^t = \hat{A}^t$, $\forall t \in [T]$
    \EndFor
    \Statex \textbf{Soil Moisture Prediction}
    \State Pass the reconstructed graph from the last time step of training, along with attribute matrices $X^{t-w} \cdots X^{t-1}$ to the trained GNN+RNN model to predict soil moisture at time $t$, $t = T+1,\cdots,T+L$.
	\end{algorithmic}
  \end{algorithm} 

\section{Experimental Section}\label{sec:exp}

\subsection{Datasets and Experimental Setup}
In this section, we summarize the datasets, baselines and the setup used for experiments in Section \ref{sec:expSetup} in the appendix. Obtaining ground truth soil moisture is expensive due to the cost of physical sensors and most of the existing works from agriculture use their private and often interpolated datasets. So, we have curated two soil moisture datasets and the associated features from multiple sources. We refer them as \textbf{Spain} and \textbf{USA}. We use 6 input features to each location for each time step for Spain and 15 input features for USA. For Spain, they are NDVI obtained from MODIS, SAR back scattering coefficients VV and VH from Sentinel 1, weather parameters consisting of temperature, relative humidity and precipitation. Locally sensed soil moisture is collected from REMEDHUS \cite{sanchez2012validation} for Spain. For USA, the features (soil temperature, weather data, etc.) and soil moisture data are obtained from the SCAN network.

We have used a diverse set of baselines algorithms like \textbf{SVR} \cite{drucker1997support} and \textbf{SVR-Shared} (for all the Shared models the parameters being shared for all the stations in a dataset), \textbf{Spatial-SVR} where spatial information in SVR is used by concatenating a node's feature with the features of its nearest k neighbors, \textbf{ARIMA} \cite{hannan1982recursive}, \textbf{RNN} \cite{chung2014empirical} and \textbf{RNN-Shared}, popular spatio-temporal GNN algorithms like \textbf{DCRNN} \cite{li2018diffusion} and \textbf{EvolveGCN} \cite{pareja2020evolvegcn}.

Hyperparameters are tuned by using the validation set and test results are reported at the best validation epoch. The embedding dimension is set to 10 for Spain and 20 for USA. We run the algorithms for 2000 epochs on Spain and 500 epochs on USA. We used three different metrics during the test period of each time series. They are Root Mean Square Error (RMSE), Symmetric Mean Absolute Percentage Error (SMAPE) and Correlation Coefficient.

\begin{table*}[ht]
\centering
\resizebox{\linewidth}{!}{
\begin{tabular}{c|ccc|ccc}
    \toprule
    
     & \multicolumn{3}{c}{Spain} &\multicolumn{3}{c}{USA}\\
    \cmidrule{2-7}
    Algorithms & RMSE ($\downarrow$) & SMAPE \% ($\downarrow$) & Correlation ($\uparrow$)  & RMSE ($\downarrow$) & SMAPE \% ($\downarrow$) & Correlation ($\uparrow$) \\
    % \cline{1-10}
    \midrule
    SVR-Shared  & 0.061 $\pm$ 0.004  & 33.654 $\pm$ 1.236  &  0.020 $\pm$ 0.001    & 12.332 $\pm$ 0.848 & 39.013 $\pm$ 3.395 & 0.287 $\pm$ 0.031  \\
    SVR & 0.052 $\pm$ 0.005 & 23.840 $\pm$ 2.453   &  0.083 $\pm$ 0.002   & 10.461 $\pm$ 0.718 & 27.377 $\pm$ 3.314  & 0.354 $\pm$ 0.031 \\
    Spatial-SVR & 0.052 $\pm$ 0.007 & 23.752 $\pm$ 1.562 & 0.150 $\pm$ 0.013 & 10.245 $\pm$ 1.567 & 26.710 $\pm$ 4.049 & 0.351 $\pm$ 0.049\\
     ARIMA  & 0.041 $\pm$ 0.008 & 19.002 $\pm$ 0.872 & 0.010 $\pm$ 0.001 & 9.290 $\pm$ 1.110  & 27.785 $\pm$ 1.306 & 0.159 $\pm$ 0.016    \\
     RNN-Shared & 0.039  $\pm$ 0.003  & 23.443 $\pm$ 1.996  & 0.585  $\pm$ 0.048  & 7.814 $\pm$ 0.442  &  30.880 $\pm$ 2.355  &  0.191 $\pm$ 0.041 \\ 
     RNN & 0.039 $\pm$ 0.003 & 21.722 $\pm$ 1.385 & 0.529 $\pm$ 0.049 & 7.338 $\pm$ 0.346  & 27.833 $\pm$ 1.812 & 0.273 $\pm$ 0.046 \\
     DCRNN & 0.061 $\pm$ 0.003  & 31.832 $\pm$ 2.363 & 0.588 $\pm$ 0.062  & 8.161 $\pm$ 0.762  &  25.213 $\pm$ 1.762  & 0.393 $\pm$ 0.015  \\
     EvolveGCN &0.061 $\pm$ 0.004 & 31.789 $\pm$ 2.554 & 0.731 $\pm$ 0.022 & 8.526  $\pm$ 0.519  & 27.717 $\pm$ 1.919  & 0.415 $\pm$ 0.038  \\
     \midrule
      DGLR (Shared) & 0.037 $\pm$ 0.003 & 17.533 $\pm$ 1.000  & 0.752 $\pm$ 0.020  & 7.526 $\pm$ 0.519  & 20.700 $\pm$ 1.210  & 0.493 $\pm$ 0.031 \\
      DGLR (w/o SL) & 0.035 $\pm$ 0.003 & 15.599 $\pm$ 0.940  & 0.753 $\pm$ 0.017  & 6.970 $\pm$ 0.356  & 18.360 $\pm$ 0.987  & 0.517 $\pm$ 0.023   \\
     DGLR (w/o Sm) &0.033 $\pm$ 0.002 & 19.835 $\pm$ 1.005  &  0.759 $\pm$ 0.018 & 6.721 $\pm$ 0.176  & 17.991 $\pm$ 1.211 &  0.533 $\pm$ 0.013 \\
     DGLR (full model) & \textbf{0.031 $\pm$ 0.001} & \textbf{15.583 $\pm$ 1.008} & \textbf{0.764 $\pm$ 0.017} & \textbf{6.454 $\pm$ 0.111} & \textbf{17.002 $\pm$ 0.987} & \textbf{0.566 $\pm$ 0.008}\\
    \bottomrule
\end{tabular}
}
\caption{Performance of soil moisture forecasting in test interval of different datasets. DGLR, as it exploits both spatial and temporal nature of the problem along with learns the graph structure, is able to perform better in most of the cases.}
\label{tab:predResults}
\end{table*}

\begin{table}[ht]
\centering
\resizebox{\linewidth}{!}{
\begin{tabular}{c|ccc}
    \toprule
     Missing
     & \multicolumn{3}{c}{Spain}\\
    \cmidrule{2-4}
  SM \% & RMSE ($\downarrow$) & SMAPE \% ($\downarrow$) & Correlation ($\uparrow$)  \\
    %\cline{1-10}
    \midrule
    0\% & 0.031 $\pm$ 0.001  & 15.583 $\pm$ 1.008  & 0.764 $\pm$ 0.017 \\
    10\% & 0.035 $\pm$ 0.004 & 18.121 $\pm$ 1.430 & 0.750 $\pm$ 0.033 \\
     20\% & 0.039 $\pm$ 0.002 & 18.864 $\pm$ 1.516 & 0.738 $\pm$ 0.027\\
     30\% & 0.039 $\pm$ 0.006 & 18.962 $\pm$ 1.112 & 0.661 $\pm$ 0.047\\ 
     
    \bottomrule
\end{tabular}
}
\caption{Test set performance of DGLR with different proportions of missing soil moisture values in the training set.}
\label{tab:sensitiSM}
\end{table}

\begin{table}[ht]
\centering
\resizebox{\linewidth}{!}{
\begin{tabular}{c|ccc}
    \toprule
     Distance 
     & \multicolumn{3}{c}{Spain}\\
    \cmidrule{2-4}
  Threshold (km) & RMSE ($\downarrow$) & SMAPE \% ($\downarrow$) & Correlation ($\uparrow$)  \\
    %\cline{1-10}
    \midrule
    8 & 0.034 $\pm$ 0.002 & 16.418 $\pm$ 0.810 & 0.752 $\pm$ 0.042 \\
    12 & 0.032 $\pm$ 0.003   & 16.441 $\pm$ 1.106  & 0.752 $\pm$ 0.028 \\
    16 & 0.031 $\pm$ 0.001 & 15.583 $\pm$ 1.008  & 0.764 $\pm$ 0.017  \\
     20 & 0.032 $\pm$ 0.004 & 16.556 $\pm$ 1.099  & 0.754 $\pm$0.054 \\
     
    \bottomrule
\end{tabular}
}
\caption{Test set performance of DGLR with different distance thresholds in initialising the graph structure.}
\label{tab:sensitiDis}
\end{table}
\subsection{Results and Analysis}
We run each algorithm 10 times on each dataset and reported the average metrics over all the locations. Table \ref{tab:predResults} shows the performance of all the baselines, along with DGLR and its variants. First, we can see that algorithms with non-shared parameters across the locations in a dataset performs better than their counterparts with shared parameters. This is because soil moisture patterns in a dataset varies significantly from one location to other. Next, the performance of dynamic GNN based baselines like DCRNN and EvolveGCN are not always better than a deep sequence modeling technique like RNN. This shows that using spatial information (via graph) may not always lead to a better performance, especially when the initially constructed graph is noisy. Finally, we find that DGLR is able to achieve  better performance than all the baseline algorithms considered.

Comparing the performance of DGLR with its variants, we can see the importance of different components of DGLR such as non-shared parameters of the GRU layers across locations, the importance of learning the graph structure and the role of feature and target smoothness regularizers.

\subsection{Sensitivity to Missing Soil Moisture Values}\label{sec:sensiSM}
The datasets that we used here do not have any missing soil moisture values. But for real-world applications, it is possible to have missing soil moisture values in between due to problem in the local sensors or in the communication networks. DGLR, being a semi-supervised approach due to use of graph neural networks, can be trained on labeled and unlabeled data points together. To see the impact of missing ground truth values in the training, we randomly remove $p\%$ soil moisture values from the training data, where $p \in \{10,20,30\}$. Table \ref{tab:sensitiSM} shows the test set performance of DGLR with different proportions of missing soil moisture values in the training. Please note that we still use the input features from those locations with missing soil moisture values and they do participate in the message passing framework of GNN. From Table \ref{tab:sensitiSM}, the performance of DGLR drops very slowly with more amount of missing soil moisture values in the training. In fact, its performance with 20\% missing values is still better than most of the baselines in Table \ref{tab:predResults} with no missing values.
So, it is clear that DGLR is quite robust in nature, and thus can be applicable for real-world soil moisture prediction.
\begin{figure}
         \begin{subfigure}[b]{0.49\linewidth}
        \centering
        \includegraphics[width=\linewidth]{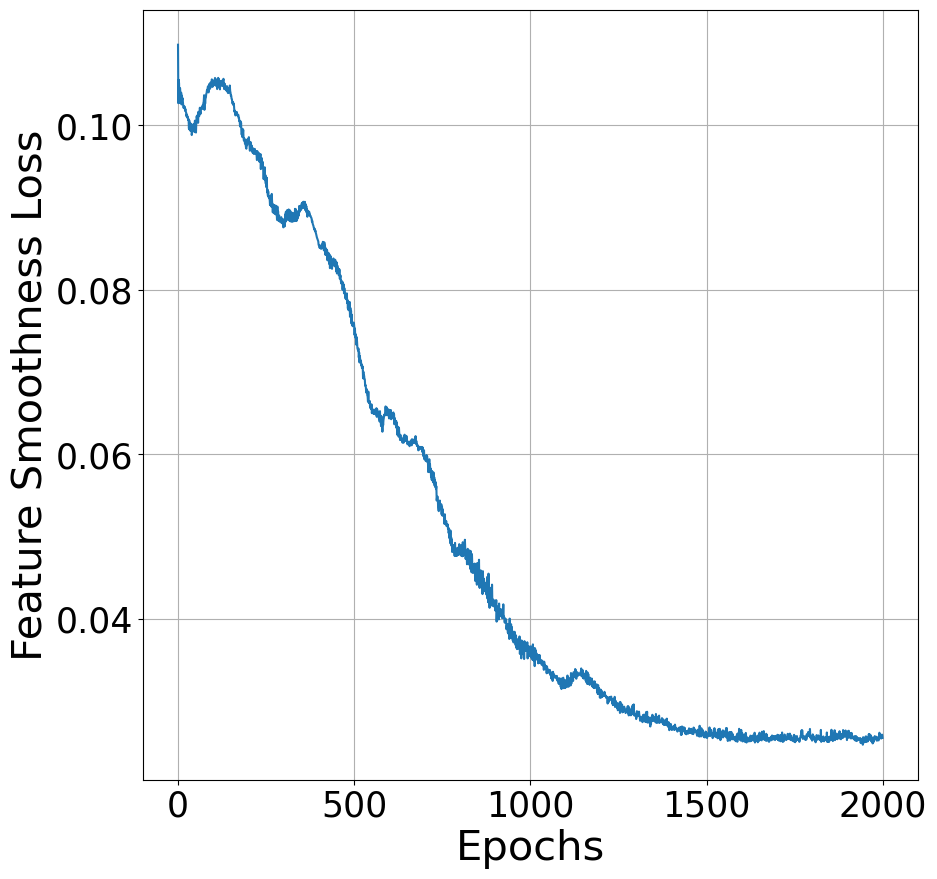}
        \vspace*{1mm}
        \caption{Feature Smoothness}
      \label{fig:featuresmoothnessloss}
    \end{subfigure}
    \begin{subfigure}[b]{0.49\linewidth}
         \centering
         \includegraphics[width=\linewidth]{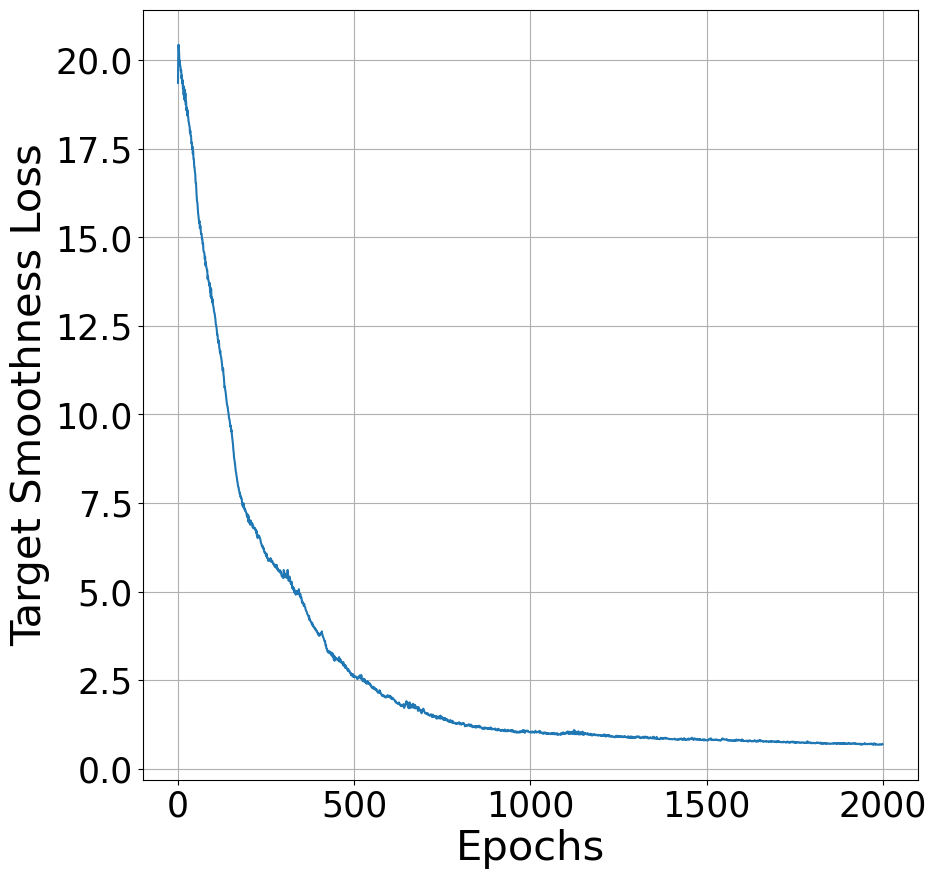}
         \vspace*{1mm}
         \caption{Target Smoothness}
         \label{fig:targetsmoothnessloss}
     \end{subfigure}
    \caption{\ref{fig:featuresmoothnessloss} and \ref{fig:targetsmoothnessloss} show loss components to depict the evolution of the graph structure of Spain over the iterations of DGLR (best seen in color).}
    \label{fig:graphEvolution}
\end{figure}

\subsection{Sensitivity to Initially Constructed Graph Structure}\label{sec:sensiGS}
We use a simple heuristic to construct an initial graph in Section \ref{sec:initial} on which we apply the proposed dynamic GNN to forecast soil moisture and update the graph structures over the iterations. In this section, we check the sensitivity of DGLR on different initially constructed graph. To do this, we vary the threshold parameter $\theta$ (Section \ref{sec:initial}). As the value of $\theta$ increases, nodes with larger distance will get connected in the initial graph, increasing the overall density. Interestingly, Table \ref{tab:sensitiDis} shows that the performance of DGLR does not change much with different values of $\theta$. This happens because DGLR also learns the optimal graph structure for soil moisture prediction. Thus, it is less sensitive to the heuristic-based initially constructed graph.

\subsection{Evolution of Graph Structure}\label{sec:graphEvo}
One key component of DGLR is to learn and update the graph structure as it predicts the soil moisture values (Section \ref{sec:graphLearning}). 
%Here, we show the values of Feature Smoothness (FS) loss (Eq. \ref{eq:FS}) and Target Smoothness (TS) loss (Eq. \ref{eq:TS}) over different iterations (Step 3 of Alg. \ref{alg:DGLR}) of DGLR. 
Feature Smoothness (FS) loss (Eq. \ref{eq:FS}) measures if nodes having dissimilar features are connected. Target Smoothness (TS) loss (Eq. \ref{eq:TS}) measures if nodes having dissimilar soil moisture values are connected. For Spain, these losses over different iterations (Step 3 of Alg. \ref{alg:DGLR}) of DGLR are shown in Figure \ref{fig:graphEvolution}. Both FS loss and TS loss decrease significantly after the initial iterations and stabilize during the end of training. Thus, DGLR is able to connect nodes with similar features and similar soil moisture values by edges over the training such a way that it also helps to predict the soil moisture better.

\section{Discussion and Conclusion}\label{sec:dis}
In this work, we have addressed the problem of soil moisture modeling and prediction which is of immense practical importance. We propose a semi-supervised dynamic graph neural network which also learns the temporal graph structures iteratively to predict soil moisture. Our solution is robust in nature with respect to missing ground truth soil moisture values and also able to achieve state-of-the-art performance for data driven soil moisture prediction on real-world soil moisture datasets. 
We hope that our solution along with the publicly available source code and datasets would help further AI research in this direction. In future, we want to apply this solution for other types of spatio-temporal data with different types of domain specific challenges. We will also check the impact of our solution in broader sustainability aspects such as improving production and quality of crops, preserving natural resources, etc.

\bibliographystyle{named}
\bibliography{ijcai22}

\appendix

\section{Related and Background Works}\label{sec:rw}
We briefly cover some background on soil moisture estimation and graph representation learning, and cite existing GNNs for dynamic graphs in this section.
Conventional machine learning approaches have been applied with diverse set of features for soil moisture estimation \cite{das2018evaluation}. Soil moisture is directly related to weather parameters such as temperature, precipitation and relative humidity. It also depends on the agricultural use of the land. Normalized Difference Vegetation Index (NDVI) \cite{pettorelli2013normalized} is a popularly used metric to measure the greenness of a field. NDVI can also be estimated using satellite data from sources like MODIS and Sentinel-2 \cite{d2013experimental}. NDVI can provide useful information for soil moisture prediction. Besides, there are other sources such as soil type, landscape information and agricultural practices such as irrigation scheduling which have direct impact on moisture contained in the soil. But they are often difficult to obtain for real-life applications. 

Next, the goal of graph representation learning is to obtain vector representation of different components of the graph (such as nodes) \cite{hamilton2017representation}. Such vector representations can be directly fed to machine learning algorithms to facilitate graph mining tasks. There exist random walk \cite{perozzi2014deepwalk}, matrix factorization \cite{bandyopadhyay2019outlier} and deep neural network \cite{wang2016structural} based algorithms for graph representation learning. However, graph neural networks (GNNs) are able to achieve significant attention in last few years due to their efficiency in graph representation and performance on downstream tasks. Most of the GNNs can be expressed in the form of the message passing network given as \cite{gilmer2017neural}: 
\begin{equation}
    h_v^l = COM^l\Bigg( \Big\{ h_v^{l-1}, AGG^k \big( \{h_{v'}^{l-1} : v' \in \mathcal{N}_G(v) \} \big) \Big\}\Bigg),
\end{equation}
where, $h_v^l$ is the representation of node $v$ of graph $G$ in $l$-th layer of the GNN. The function $AGG$ (Aggregate) considers representation of the neighboring nodes of $v$ from the $(l-1)$th layer of the GNN and maps them into a single vector representation. As neighbors of a node do not have any ordering in a graph and the number of neighbors can vary for different nodes, $AGG$ function needs to be permutation invariant and should be able to handle different number of nodes as input. Then, $COM$ (Combine) function uses the node representation of $v$th node from $(l-1)$th layer of GNN and the aggregated information from the neighbors to obtain an updated representation of the node $v$.

Graph neural networks are also proposed for dynamic graphs in recent years. Traffic data forecasting can naturally be posed as a prediction problem on dynamic graphs \cite{li2018diffusion}. Temporal convolution blocks along with different types of recurrent networks are used for traffic data forecasting using GNNs \cite{guo2019attention,yu2018spatio}. Using the gated
attention networks as a building block, graph gated recurrent unit is proposed to address the traffic speed forecasting problem by \cite{DBLP:conf/uai/ZhangSXMKY18}. Deep learning has been used for traffic management and crowd safety as an application of spatio-temporal data modeling \cite{zhang2017deep}. Traffic forecasting problem is converted to a node-wise graph and an edge-wise graph, and then a bi-component graph convolution is applied in \cite{DBLP:conf/aaai/ChenCXCGF20}. A space-time graph neural network is proposed by \cite{nicolicioiu2019recurrent} where both nodes and edges have dedicated neural networks for processing information. A GNN architecture, Graph WaveNet, for spatial-temporal graph modeling to handle long sequences is proposed by \cite{wu2019graph}. Recurrent neural network structures that operate over a graph convolution layer to handle dynamic graphs are proposed by \cite{sankar2020dysat,pareja2020evolvegcn}.
The problem of ride-hailing demand forecasting is encoded into multiple graphs and multi-graph convolution is used in \cite{geng2019spatiotemporal}. An approach for automated neural
architecture search for spatio-temporal graphs with the application to urban traffic prediction is done in \cite{pan2021autostg}.
However, the problem of soil moisture estimation (or similar metrics for precision agricultural applications) is quite different because of the dependency to multiple external factors and also uncertainty in the graph structure to capture spatial correlation. 
Recently, research is conducted on graph structure learning in the framework of graph neural networks \cite{franceschi2019learning,elinas2020variational,zhao2021data,shang2020discrete,fatemi2021slaps,zhu2021deep}. But these studies are mostly limited to learning a static graph structure. Where for our application, a dynamic graph is needed since the input data changes over time. Moreover, they often parameterize the entries of the adjacency matrix, thus making the optimization combinatorial in nature.

\section{Solution Approach}
\begin{figure}
    \centering
    \includegraphics[width=\linewidth]{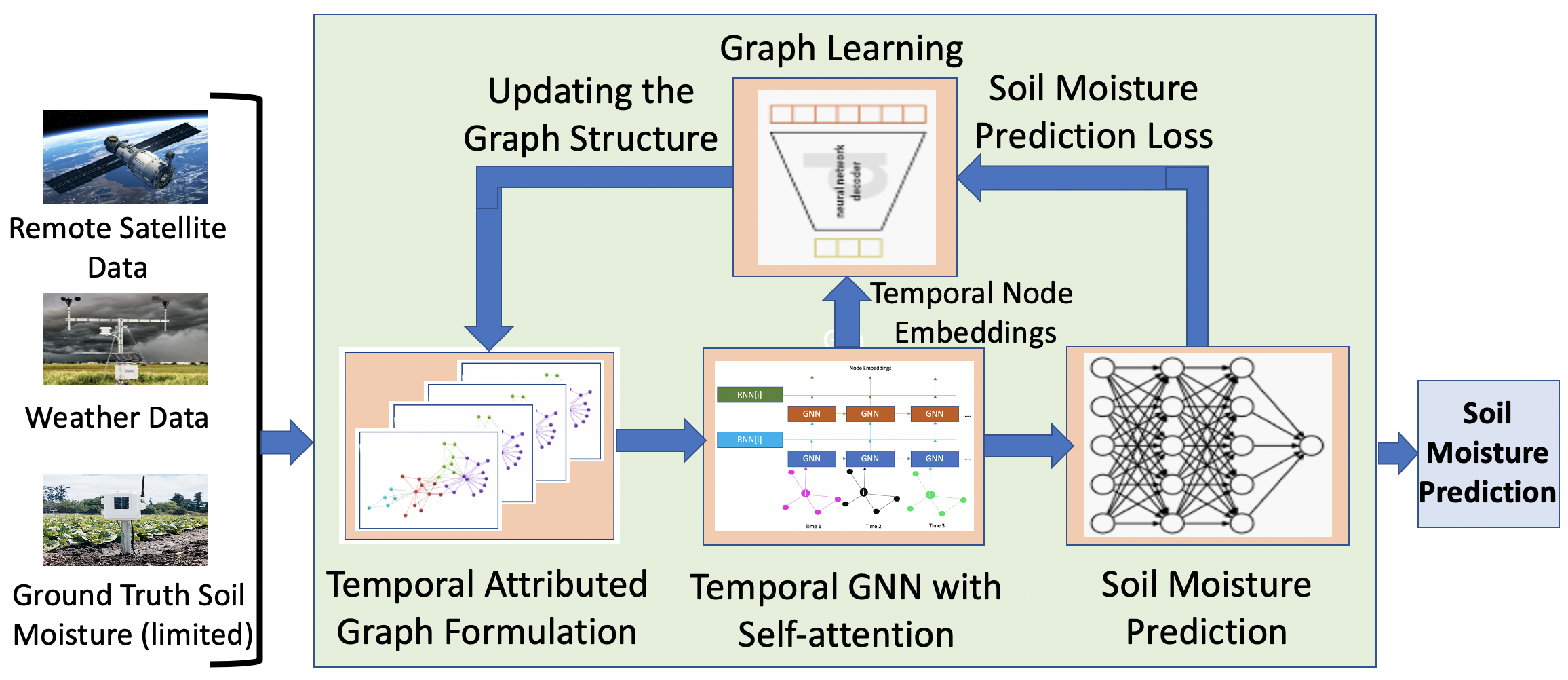}
    \caption{Solution Approach for Soil Moisture Modeling}
  \label{fig:soln}
\end{figure}

\begin{figure}
    \centering
    \includegraphics[width=\linewidth]{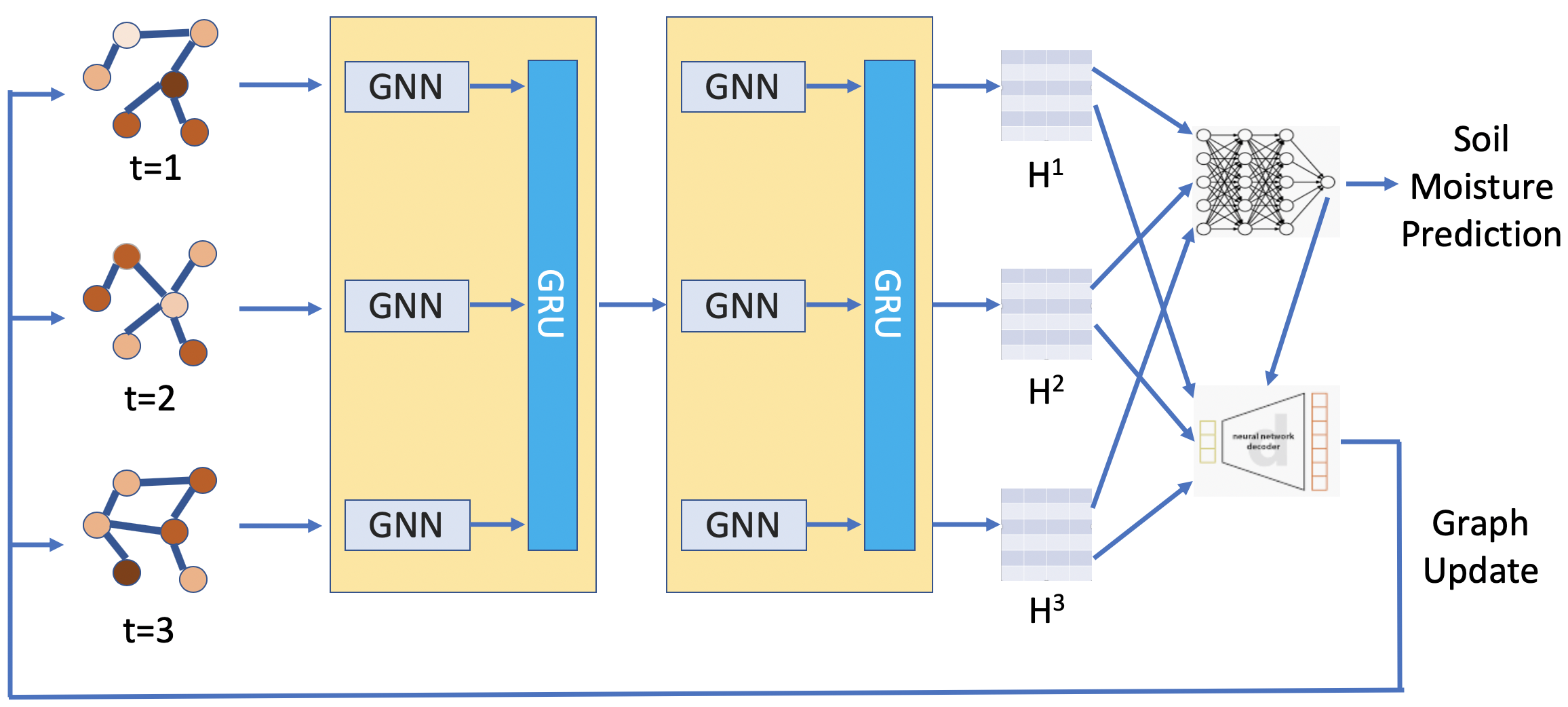}
    \caption{Neural Architecture of DGLR}
  \label{fig:neuraldglr}
\end{figure}

Figure \ref{fig:soln} gives a high level overview of our solution approach, as explained in details in the main paper. Different notations used in the paper are summarized in Table \ref{tab:notations}. We have included the detailed neural architecture of DGLR in Figure \ref{fig:neuraldglr}. Given the initially constructed graphs from the input locations and features, DGLR computes the embedding of each node in each time step of the given dynamic graph. DGLR use those node embeddings to predict soil moisture and reconstruct the dynamic graph by minimizing the loss on various components of the solution (Equation \ref{eq:totalLoss}).

\begin{table}
\centering
\resizebox{\linewidth}{!}{%
\begin{tabular}{*6c}
	\toprule
	\sffamily{Notations} & Explanations\\
%    \sffamily{} & & & & Words & Distribution & links \\
    \hline
	\midrule
	$i,j \in \{1,2,\cdots,N\}=[N]$ & Indices over locations (nodes) \\
	$d_{ij} \in \mathbb{R}_+$ & Geographic distance between $i$ and $j$\\
	$t \in \{1,2,\cdots,T\}=[T]$ & Indices over time steps \\
	$N_{tr}^t \in [N]$ & Locations with soil moisture values at time $t$\\
    $G^t=(V,E^t,X^t)$ & The graph at time $t \in [T]$ \\
    $x_i^t \in \mathbb{R}^D$ & Input node features for $i$th node at time $t$ \\
    $h_i^t \in \mathbb{R}^{K}$ & Node embedding of $i$ th node in $G^t$ \\
    $s_i^t \in \mathbb{R}$ & Actual soil moisture for $i$th station at time $t$\\
    $\hat{s}_i^t$ & Predicted soil moisture\\
    $\theta$ & Distance threshold for initial graph construction \\
    $A^t \in \mathbb{R}^{N \times N}$ & Initially constructed adjacency matrix \\
    $\hat{A}^t = H^t {H^t}^{Trans}$ & Reconstructed adjacency matrix \\
    $W_{G}$ & Parameter of the GNN for the graphs\\
    $\Theta_{R} = \{\Theta_{R,i}:\forall i\}$ & Parameters of the RNNs for all the nodes\\
    \bottomrule
	\end{tabular}
	}
\caption{Different notations used in the paper}
\label{tab:notations}
\end{table}

\section{Definitions of the Metrics Used}
Formal definition of RMSE and SMAPE are given below. They are both used to measure the error in time series prediction (forecasting),
\begin{flalign}
    RMSE = \sqrt{ \frac{(\sum\limits_{t=T+1}^L s_i^t-\hat{s}_i^t)^2}{L} }, \\
    SMAPE = \frac{100\%}{L}\sum\limits_{t=T+1}^L \frac{|s_i^t-\hat{s}_i^t|}{(s_i^t+\hat{s}_i^t)/2},
\end{flalign}
where we assumed the test interval to be in $[T+1,\cdots,L]$.
Pearson correlation coefficient is calculated as
\begin{flalign}
    corr = \frac{\sum\limits_{t=T+1}^L (s_i^t - \bar{s}_i) (\hat{s}_i^t - \bar{\hat{s}}_i)}
    {\sqrt{\sum\limits_{t=T+1}^L (s_i^t - \bar{s}_i)^2} \sqrt{\sum\limits_{t=T+1}^L(\hat{s}_i^t - \bar{\hat{s}}_i)^2}},
\end{flalign}
where $\bar{s}_i = \frac{\sum\limits_{t=T+1}^L s_i^t}{L}$ and $\bar{\hat{s}}_i = \frac{\sum\limits_{t=T+1}^L \hat{s}_i^t}{L}$.

The above metrics are calculated for each station $i \in [N]$ in a dataset and results are reported in Tables 2 and 3 of the main paper by taking the average (along with standard deviations) over all the stations, and over 10 different runs.

\section{Experimental Section}\label{sec:exp}
In this section, we discuss the datasets and baseline algorithms, and analyze the results for soil moisture prediction.

\begin{figure}
         \begin{subfigure}[b]{0.49\linewidth}
        \centering
        \includegraphics[width=\linewidth]{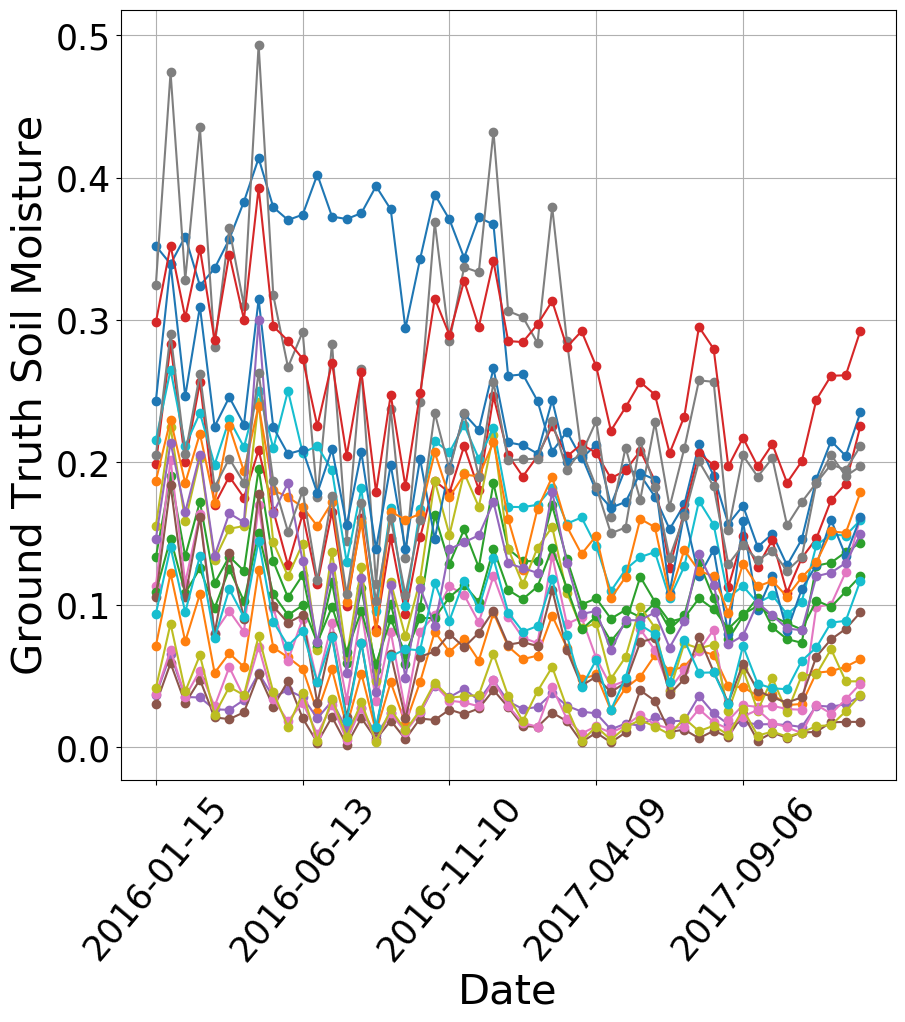}
        \vspace*{1mm}
    \caption{Soil Moisture in Spain}
      \label{fig:spain}
    \end{subfigure}
    \begin{subfigure}[b]{0.49\linewidth}
         \centering
         \includegraphics[width=\linewidth]{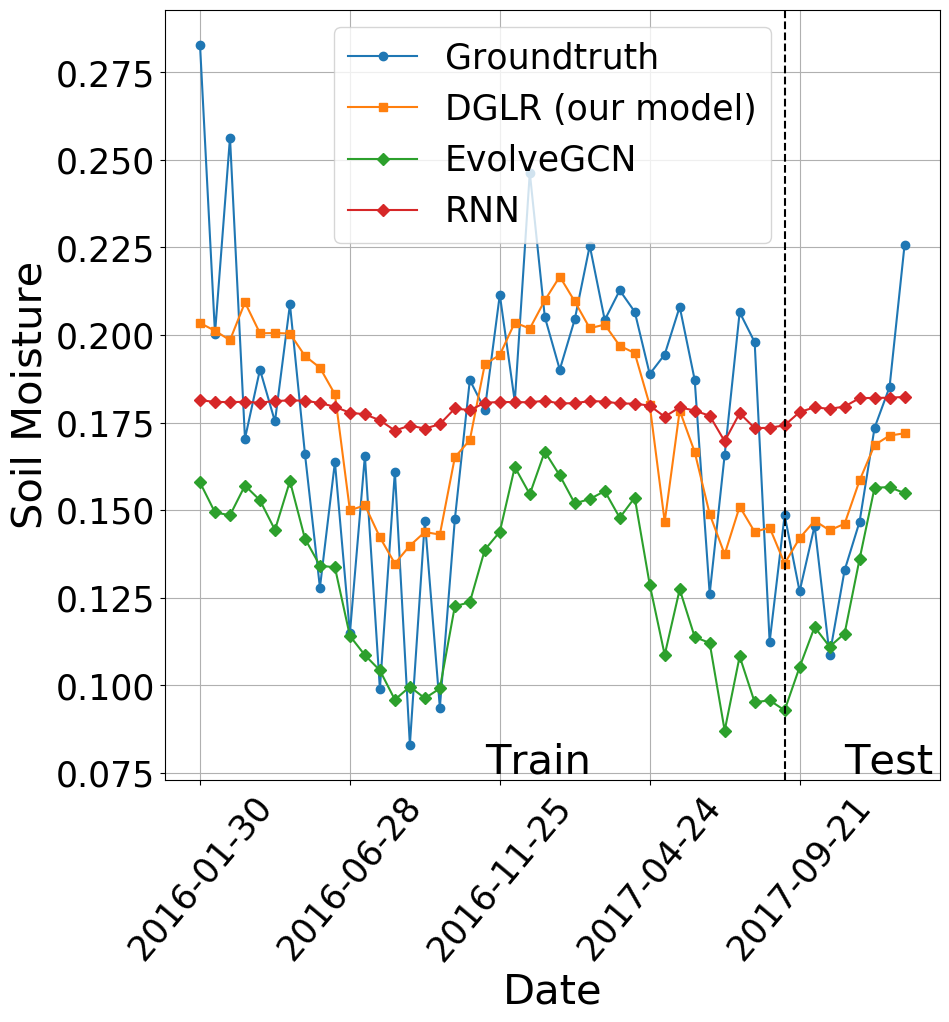}
         \vspace*{1mm}
          \caption{Prediction for Spain}
      \label{fig:pred}
     \end{subfigure}
    \caption{\ref{fig:spain} shows all the ground truth soil moisture (SM) time series from Spain. \ref{fig:pred} shows forecasted SM by different algorithms on a randomly selected station from Spain (best seen in color).}
    \label{fig:datasetsAndLosses}
\end{figure}

\subsection{Datasets Used}\label{sec:datasets}
There is no standard benchmarking dataset available for soil moisture prediction. Obtaining ground truth soil moisture is very difficult as one needs to install a physical soil moisture sensor at each location. Besides, most of the existing works from agriculture use their private and often interpolated datasets. However, we have collected following two soil moisture datasets. We make both the source code of DGLR and the datasets publicly available.

\textbf{Spain}: The dataset consists of 20 soil moisture stations from North-Western Spain, for the years 2016-2017, with a temporal resolution of 15 days based on the availability of the features. It has a total of 49 time steps. We use the data from the fist 40 time steps for training and validation, and last 9 time steps for testing. This dataset has six temporal features such as NDVI obtained from MODIS (\url{https://modis.gsfc.nasa.gov/}), SAR back scattering coefficients VV and VH from Sentinel 1
(\url{https://sentinel.esa.int/}), weather parameters consisting of temperature, relative humidity and precipitation (\url{https://www.ibm.com/weather}). Locally sensed soil moisture is collected from REMEDHUS \cite{sanchez2012validation}. We have plotted the ground truth soil moisture time series for each station in Spain in Figure \ref{fig:spain}. 

\textbf{USA}: The dataset consists of 68 soil moisture stations from USA, for the years 2018-2019, with a daily resolution based on the availability of the features. It has a total of 731 time steps. We use the data from the fist 518 time steps for training and validation, and last 212 time steps for testing. This dataset has 15 temporal features (such as different weather and soil parameters) for each location from SCAN network (\url{https://www.wcc.nrcs.usda.gov/scan/}). 

\subsection{Baseline Algorithms}
We have used a diverse set of baseline algorithms to compare with the performance of DGLR, as follows.

\textbf{SVR} and \textbf{SVR-Shared} \cite{drucker1997support}: We train Support Vector Regression (SVR) to model soil moisture with respect to the features, independently for each station. For SVR-Shared, we use the same SVR model (parameters being shared) for all the stations in a dataset.

\textbf{Spatial-SVR}: To use spatial information in SVR (non-shared), we created this baseline by concatenating a node's feature with the features of its nearest k neighbors (in order, based on geographic distance). We set k = 4 for Spain and k = 10 for USA, based on average degree of the nodes in the initially formed graph in Section \ref{sec:initial}.

\textbf{ARIMA} \cite{hannan1982recursive}: We use ARIMA with Kalman filter which is used in time series forecasting.

\textbf{RNN} and \textbf{RNN-Shared} \cite{chung2014empirical}: A set of two RNN (GRU) layers is used on the input features for each location over time. Models are trained independently across the locations. For RNN-Shared, we train only a single GRU model (having two layers) for all the locations.

\textbf{DCRNN} \cite{li2018diffusion}: DRCNN is a popular spatio-temporal GNN used for traffic forecasting problem. It captures the spatial dependency using bidirectional random walks on the graph, and the temporal dependency using the encoder-decoder architecture with scheduled sampling. DCRNN does not learn the graph structure. So we use it on the initially constructed graph at Section \ref{sec:initial}. We also use the Unicov version of DCRNN as the graph is symmetric.

\textbf{EvolveGCN} \cite{pareja2020evolvegcn}: This is a recently proposed dynamic graph neural network which uses a RNN on the parameter matrices of GNN, thus captures both graph structure and the temporal nature of the problem. Similar to DCRNN, we use EvolveGCN on the initially constructed graph in Section \ref{sec:initial}.

\subsection{Experimental Setup and Hyperparameters} \label{sec:expSetup}
We have run all the experiments on a CPU with 2.6GHz 6‑core Intel Core i7. We conduct extensive hyperparameter tuning for all the baseline algorithms on a small part of the training set (i.e., validation set).
Hyperparameters are tuned by using the validation set and test results are reported at the best validation epoch. The threshold value to connect edges in the initial graph is 16 km for Spain and 1000 km for USA (threshold is high because of large distances between stations). We use grid search to fix the values of the hyperparameters. We keep the grid search range of the common hyperparameters same for the baselines and our proposed algorithms to ensure a fair comparison. For example, grid search for window length is in range $[1,4]$ for both the datasets and the learning rate is in range $[0.0001,0.001,0.01]$ for Spain and $[0.01,1]$ for USA, for all the algorithms. The embedding dimension is set to 10 for Spain and 20 for USA. For DGLR and its variants, we fix the values of the weights (for example, $\alpha_1$, $\alpha_2$, $\alpha_3$ and $\alpha_4$) in such a way that contributions from the respective components of the cost function become similar at the first iteration of the algorithm. We run the algorithms for 2000 epochs on Spain and 500 epochs on USA.

We used three different metrics \cite{tofallis2015better} to analyze the performance of the algorithms during the test period of each time series. They are Root Mean Square Error (RMSE), Symmetric Mean Absolute Percentage Error (SMAPE) and Correlation Coefficient. RMSE and SMAPE both measure the error in time series prediction. So lesser the value of them better is the quality. For correlation, more is the value better is the quality. RMSE is not bounded in nature, whereas SMAPE is in the range of $[0\%,100\%]$ and correlation is in $[-1,+1]$.

Figure \ref{fig:pred} shows the ground truth soil moisture from a station in Spain and predicted soil moisture by DGLR and some baselines. It is quite evident that DGLR is able to match both trend and local patterns of the soil moisture, both during training and test intervals, compared to other baselines.

\end{document}